\documentclass[10pt,twocolumn,letterpaper]{article}

\usepackage[pagenumbers]{cvpr} 

\definecolor{royalblue}{HTML}{4169E1}
\definecolor{royalred}{HTML}{C71A1A}
\definecolor{cvprblue}{rgb}{0.21,0.49,0.74}
\usepackage[pagebackref,breaklinks,colorlinks,allcolors=cvprblue]{hyperref}
\usepackage{soul}
\usepackage{array}

\def\paperID{26} 
\def\confName{CVPR}
\def\confYear{2026}

\title{I Walk the Line: Examining the Role of Gestalt Continuity in Object Binding for Vision Transformers}

\author{Alexa R. Tartaglini\\
Stanford University\\
{\tt\small alexart@stanford.edu}
\and
Michael A. Lepori\\
Brown University\\
{\tt\small michael\_lepori@brown.edu}
}

\begin{document}
\maketitle

\begin{abstract}
Object binding is a foundational process in visual cognition, during which low-level perceptual features are joined into object representations. Binding has been considered a fundamental challenge for neural networks, and a major milestone on the way to artificial models with flexible visual intelligence. Recently, several investigations have demonstrated evidence that binding mechanisms emerge in pretrained vision models, enabling them to associate portions of an image that contain an object. The question remains: \textbf{how} are these models binding objects together?

In this work, we investigate whether vision models rely on the principle of Gestalt continuity to perform object binding, over and above other principles like similarity and proximity. Using synthetic datasets, we demonstrate that binding probes are sensitive to continuity across a wide range of pretrained vision transformers. Next, we uncover particular attention heads that track continuity, and show that these heads generalize across datasets. Finally, we ablate these attention heads, and show that they often contribute to producing representations that encode object binding.

\end{abstract}    
\vspace{-1em}
\section{Introduction}
\label{sec:intro}

\vspace{-0.5em}
Natural scenes are replete with low-level perceptual features like colors, shapes, textures, and motion cues. Despite this complexity, humans can instantly parse this information into sets of coherent objects. This ability to join low-level features together into object-level representations is known as object binding \citep{treisman1996binding}, and it has long been thought to play a fundamental role in visual cognition \citep{treisman1982illusory, von1999and, greff2020binding}.

Many performant contemporary vision models are based on the transformer architecture, which processes images as sets of fixed-size square patches \citep{dosovitskiy2020image, vaswani2017attention}. Recent work has suggested that these models have learned mechanisms for binding objects, resulting in representations indicating which image patches contain a particular object \citep{li2025does}. The question remains: \textit{how} are these models implementing object binding (i.e., how does the model know which patches to assign binding representations to)? For insight, we look to the study of human perception. 20th century Gestalt Psychology has long articulated principles that humans rely on when parsing scenes from low-level perceptual cues \citep{wertheimer2017untersuchungen, koffka2013principles}. These principles (e.g. similarity, proximity, closure, etc.) describe common low-level visual motifs that suggest structure within the visual world. In this work, we focus on the principle of \textit{continuity}: the intuitive heuristic that visual elements that form a continuous curve are likely to belong to the same object or contour.

We create controlled synthetic datasets in order to isolate the effect of continuity from two other Gestalt principles: proximity and similarity. Across a range of vision transformer models, we find that continuity routinely contributes to forming binding representations. Furthermore, we discover specific attention heads --- \textit{Gestalt continuity heads} --- that track continuous curves across image patches, and show that these naturally arise across vision models that were trained using diverse learning objectives. Finally, we analyze the impact of ablating Gestalt continuity heads on object binding, and observe that ablating these heads often selectively impacts binding representations for images that can leverage continuity for object binding.
\section{Related Work}
\label{sec:related}

\begin{figure*}[t]
  \centering
  \begin{subfigure}[b]{0.0927\linewidth}
    \includegraphics[width=\textwidth]{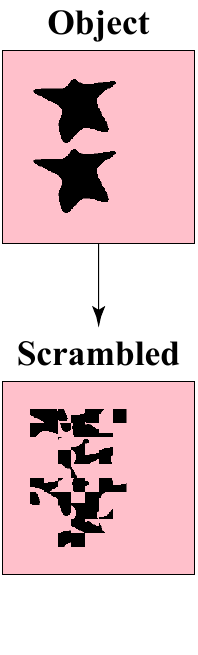}
    \caption{}
    \label{fig:probe-a}
  \end{subfigure}
  \hfill
  \begin{subfigure}[b]{0.48\linewidth}
    \includegraphics[width=\textwidth]{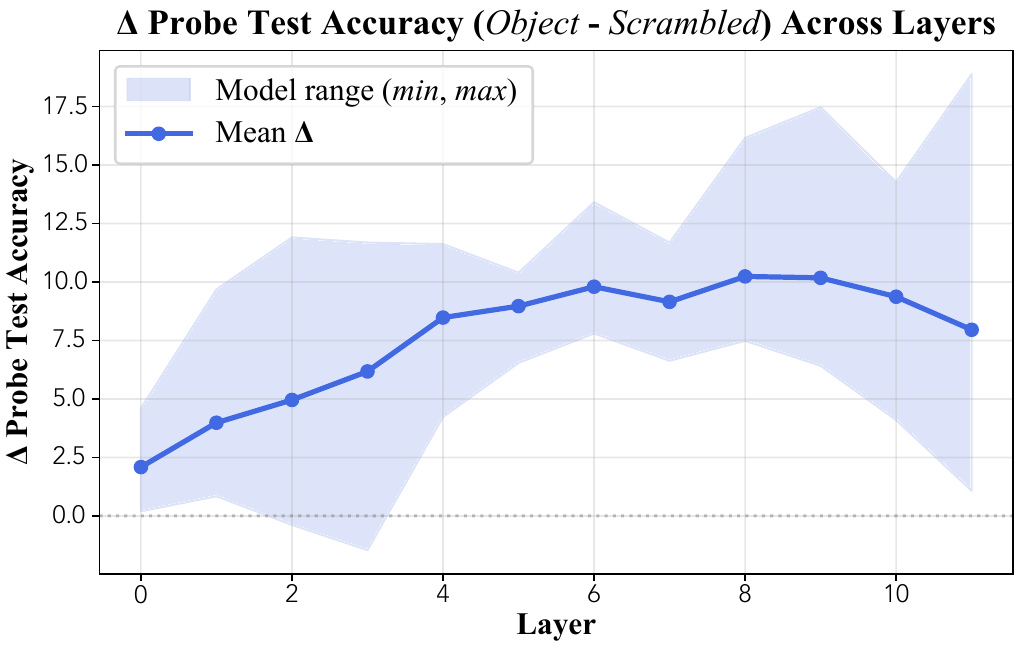}
    \caption{}
    \label{fig:probe-b}
  \end{subfigure}
  \begin{subfigure}{0.4\linewidth}
    \centering
    \small
    \begin{tabular}{>{\footnotesize}c>{\footnotesize}l>{\footnotesize}c}
        & \multicolumn{2}{c}{$\max(\text{\textbf{Probe Test Acc.}})$} \\
        \toprule
        \textbf{\small Model} & \multicolumn{1}{c}{Blobs} & \multicolumn{1}{c}{Curves} \\
        \midrule
        DINO ViT-B/16 & 91.2\% \textcolor{royalblue}{(+4.3)} & 86.9\% \textcolor{royalred}{(+10.6)} \\
        DINOv2 ViT-B/14 & 98.7\% \textcolor{royalblue}{(+1.4)} & 97.3\% \textcolor{royalred}{(+16.6)} \\
        MAE ViT-B/16 & 99.0\% \textcolor{royalblue}{(+3.7)} & 95.3\% \textcolor{royalred}{(+16.1)} \\
        ImageNet ViT-B/16 & 95.5\% \textcolor{royalblue}{(+6.1)} & 89.4\% \textcolor{royalred}{(+11.7)} \\
        ImageNet ViT-B/32 & 95.5\% \textcolor{royalblue}{(+4.6)} & 90.9\% \textcolor{royalred}{(+15.6)} \\
        CLIP ViT-B/16 & 91.3\% \textcolor{royalblue}{(+1.4)} & 89.9\% \textcolor{royalred}{(+9.4)} \\
        CLIP ViT-B/32 & 90.4\% \textcolor{royalblue}{(+3.2)} & 87.2\% \textcolor{royalred}{(+11.2)} \\
        \bottomrule
    \end{tabular}
    \vspace{3em}
    \caption{}
    \label{tab:max_probe_test_acc}
  \end{subfigure}
  \caption{\textbf{Object binding probe stimuli and results.} \textbf{(a)} An example stimulus pair from the dataset used to train and test the binding probes (Object and $\text{ Scrambled}_{\textit{location}}$). More example images from the Object and Scrambled datasets can be found in Appendix~\ref{appendix:datasets}. 
  \textbf{(b)} Difference between object binding probe test accuracy on Blobs Object stimuli vs.  Scrambled stimuli (taking the maximum probe test accuracy over $\{\text{Scrambled}_{\textit{orientation}}, \text{ Scrambled}_{\textit{location}}\}$), averaged across models. See Appendix~\ref{appendix:binding} for results on the Curves dataset. \textbf{(c)} Maximum object binding probe test accuracy for Object stimuli from the Blobs and Curves datasets across model layers. The quantities in the parentheses show the delta with maximum probe test accuracy on the Scrambled datasets; for example, (+6.1) indicates that maximum probe test accuracy on the Object dataset is 6.1 points higher than the probe test accuracy on either of the Scrambled datasets. Scrambled dataset accuracies are reported explicitly in Appendix~\ref{appendix:binding}. Overall, we find that binding probes achieve higher performance when the stimuli obey continuity, even when controlling for proximity and surface similarity of the patches. This indicates that continuity plays a role in generating binding representations across vision transformer models. See Appendix~\ref{appendix:binding} for full results.
  }
  \label{fig:probe}
\end{figure*}

\vspace{-0.5em}
The ``binding problem'' is thought to pose a particular challenge to neural network models \citep{von1999and, greff2020binding}, leading some to conclude that models may require strong architectural inductive biases to learn bound object representations \citep{locatello2020object}. However, recent work has found widespread evidence of binding mechanisms across contemporary vision \citep{li2025does, lepori2024beyond}, language \citep{fenglanguage}, and multimodal \citep{assouel2025visual} transformer models. 

Much of this work posits the existence of a ``binding-ID'' subspace in a model's intermediate activations, which can be populated with ``binding vectors'' that are applied to bind activations to one another. However, it is unclear how these binding vectors are created. The nature of this mechanism is further obscured by another line of work arguing that purely feedforward networks (including transformers) may struggle with perceptual grouping according to Gestalt principles \citep{linsley2018learning, kimdisentangling, taylong}. In contrast, we demonstrate the emergence of attention heads that track continuity across patch boundaries in a variety of transformers. We then demonstrate that these heads are implicated in producing binding vectors.
\section{Methods}
\label{sec:methods}

\begin{figure*}
  \centering
  \begin{subfigure}[t]{0.42\linewidth}
    \includegraphics[width=\textwidth]{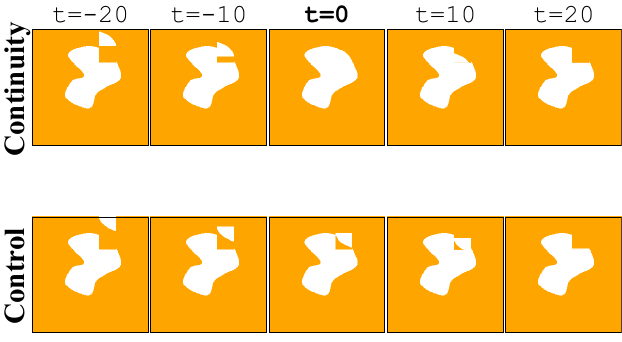}
    \caption{}
    \label{fig:continuity_head-a}
  \end{subfigure}
  \hfill
  \begin{subfigure}[t]{0.15\linewidth}
    \includegraphics[width=\textwidth]{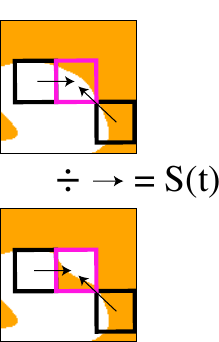}
    \caption{}
    \label{fig:continuity_head-b}
  \end{subfigure}
  \hfill
  \begin{subfigure}[t]{0.4\linewidth}
    \includegraphics[width=0.9\textwidth]{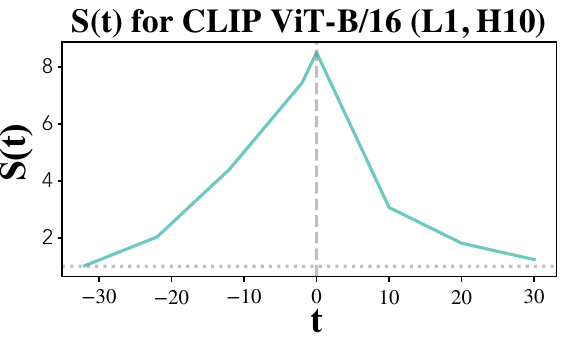}
    \caption{}
    \label{fig:continuity_head-c}
  \end{subfigure}
  \caption{\textbf{Procedure for identifying Gestalt continuity heads.} \textbf{(a)} Top row: We parametrically vary continuity by translating a target perimeter patch. At $t=0$, the patch is maximally aligned with its neighbors, creating a continuous curve. Bottom row: We create a control condition where the target patch at $t=0$ contains the same pixel information but randomly rotated as to disrupt continuity between neighboring patches. \textbf{b} At each $t$ and for each attention head, we measure attention from neighboring patches containing the curve (black squares) to the target patch (blue square). We then divide the maximum neighbor attention for the continuity condition by the same quantity for the control condition to get $S(t)$, the continuity sensitivity index. \textbf{c} $S(t)$ for an attention head from CLIP ViT-B/16 with a high continuity sensitivity index (Layer 1, Head 10). Attention peaks at the maximally-aligned time step $t=0$. See Appendix~\ref{appendix:tuning_curves} for all models.}
  \label{fig:continuity_head}
\end{figure*}

\vspace{-0.5em}
\paragraph{Models}
We study a range of vision transformers, which are pretrained on different tasks but share most of their architectural details. Specifically, we study several 12-layer self-supervised models: DINO ViT-B/16 \citep{caron2021emerging}, DINOv2 ViT-B/14 \citep{oquab2024dinov2}, and a Masked Autoencoder model (MAE ViT-B/16; \citet{he2022masked}); two versions of the CLIP model \citep{radford2021learning}: CLIP ViT-B/16 and B/32 (which have patch sizes of 16$\times$16 and 32$\times$32 pixels, respectively); and two supervised models: ImageNet ViT-B/16 and B/32 \citep{dosovitskiy2020image}.

\vspace{-1em}
\paragraph{Datasets} For our main set of experiments, we generate a synthetic dataset of filled blobs in which each image consists of a randomly colored blob on a randomly colored background. The boundary between blob and background forms a continuous closed curve whose shape we can disrupt, allowing us to probe for representations of Gestalt continuity in models. Additionally, we replicate our main results using a dataset of open curves. See Appendix~\ref{appendix:datasets} for dataset generation details and examples.

\vspace{-1em}
\paragraph{Binding Probes} We adopt the highest performing probing strategy from \citet{li2025does}, the quadratic probe. This probe takes two patch activations, $x$ and $y$, as input and computes $\sigma(xWy^T + b)$, where $W$ is constrained to be symmetric. This probe is trained with weight decay in order to encourage a low effective rank.
\section{Object Binding is Sensitive to Visual Continuity Cues}
\label{sec:binding_probes}

\vspace{-0.5em}
First, we establish that the binding representations discovered by the quadratic binding probes are sensitive to object continuity, over and above their sensitivity to object similarity and proximity. We train binding probes on a dataset comprised of images containing two copies of identical random blob shapes (see Figure~\ref{fig:probe}a, top row). The probe is trained on pairs of activations corresponding to patches belonging to these objects, and must discriminate whether the two patches belong to the same or different objects. These objects are identical in shape, color, and, texture, reducing the probe's reliance on object similarity to establish binding. We train probes for each layer of all models described in Section~\ref{sec:methods} for 20 epochs using the Adam optimizer \citep{kingma2014adam} with a learning rate of 0.001 weight decay factor of 0.01 and a stepwise learning rate scheduler with $\gamma$ = 0.2.

We then evaluate the probe on three datasets: a held out within-distribution test set of whole objects (\textbf{Object}) and two control datasets designed to disrupt continuity cues (\textbf{Scrambled}; see Figure~\ref{fig:probe}a). The $\text{\textbf{Scrambled}}_{\textit{orientation}}$ dataset randomly rotates each image patch in the Object images by $d\in\{90^\circ, 180^\circ, 270^\circ\}$. This preserves the content and position of each patch while systematically breaking the continuity of the object border. The $\text{\textbf{Scrambled}}_{\textit{location}}$ dataset randomly permutes the locations of patches that contain a particular object. This preserves the overall content and position of the \textit{set} of patches containing the object. The extent to which a model's Object probe performance exceeds its $\text{Scrambled}$ performance is indicative of its reliance on continuity cues when performing object binding (as opposed to simple patch similarity or proximity).

From Figure~\ref{fig:probe}, we see that probe performance on the Object dataset regularly outperforms performance on the Scrambled control datasets, with performance increasing from early to middle layers. However, we note that performance on the control sets far exceeds the majority-class baseline accuracy of 50.1\%(see Table~\ref{tab:max_probe_test_acc}). This indicates that models are relying on continuity \textit{in addition to} other Gestalt principles, such as proximity.

\section{Specific Attention Heads Detect Continuity}
\label{sec:gestalt_heads}

\vspace{-0.5em}
Having shown that object binding is sensitive to continuity cues in the image, we now ask whether this sensitivity can be localized to specific attention heads. We hypothesize that certain heads may act as dedicated continuity detectors, preferentially attending between neighboring patches when the visual content of each patch forms a continuous curve across patch boundaries. 

To identify these hypothesized \textit{Gestalt continuity heads}, we design an experiment that parametrically varies continuity in stimuli from our Blobs dataset. For this analysis, we use images that contain just one blob (in contrast to the datasets used to assess object binding). For each blob stimulus, we select a single perimeter patch and translate it along a trajectory such that at $t=0$ it is maximally aligned with the curve formed by the object boundary, and at $t=\pm n$ it is progressively displaced (see Figure~\ref{fig:continuity_head}a, top row). For a given attention head, we measure the maximum attention directed from neighboring curve-completing perimeter patches to the target perimeter patch at each time step $t$. We repeat this for all possible perimeter patches in a given image and average to arrive at a scalar-valued score of that head's attention to continuity.

To control for low-level content of the patches, we construct a control condition in which the target patch at each time step is randomly rotated by $d \in \{90^\circ, 270^\circ\}$ (see Figure~\ref{fig:continuity_head}a). This preserves the pixel content of the patch at $t=0$ while disrupting the continuity of the curve between the target and its neighbors. We then compute a \textit{continuity sensitivity index} $S(t)$ at each time step by dividing the maximum neighbor-to-target attention in the base condition by the corresponding quantity in the rotated condition, applying a floor of $1$ (see Figure~\ref{fig:continuity_head}b). A sensitivity index greater than $1$ indicates that the head allocates more attention when the curve is completed between neighboring patches than when it is not, despite identical pixel content at the target location; for example, a sensitivity index of $8$ means the head is $8\times$ more sensitive to continuity than mere pixel content or patch proximity. Plotting $S(t)$ across time steps produces a ``tuning curve'' \citep{dayan2005theoretical} --- heads that are sensitive to continuity exhibit a characteristic Gaussian-like profile peaking at $t=0$ where curve alignment is maximal (see Figure~\ref{fig:continuity_head}c).

Across models, we find a set of heads exhibiting high continuity sensitivity indices at $t=0$, mostly clustered within the first four layers (Appendix~\ref{appendix:continuity}). To validate these findings, we repeat the procedure on two additional stimulus sets: the Curves dataset described in Section~\ref{sec:binding_probes}, and a dataset of naturalistic objects drawn from ImageNet \citep{deng2009imagenet} (see Appendix~\ref{appendix:datasets} for details). We compute Pearson's $r$ of the continuity sensitivity index at $t=0$ over all heads between each pair of datasets (Blobs, Curves, and ImageNet). Continuity heads are highly consistent across datasets, with an average Pearson's $r$ of $0.664$ across models and dataset pairs (see Appendix~\ref{appendix:correlations} for a breakdown by model and $p$-values). Furthermore, when examining the full tuning curves across all time steps, identified continuity heads display the characteristic Gaussian-like tuning curves expected of a continuity detector: attention between neighboring patches peaks at $t=0$, where curve alignment is maximal, and decays monotonically as the target patch becomes increasingly misaligned with its neighbors (Appendix~\ref{appendix:tuning_curves}). Overall, this indicates that particular attention heads are sensitive to Gestalt continuity cues, providing at least one concrete mechanism for representing continuity across patch boundaries. We note that the concentration of continuity heads across layers roughly mirrors probe performance across layers (see Appendix~\ref{appendix:continuity}).

\section{Gestalt Heads are Causally Implicated in Binding}
\label{sec:ablation}

\begin{figure}[t]
  \centering
    \includegraphics[width=\linewidth]{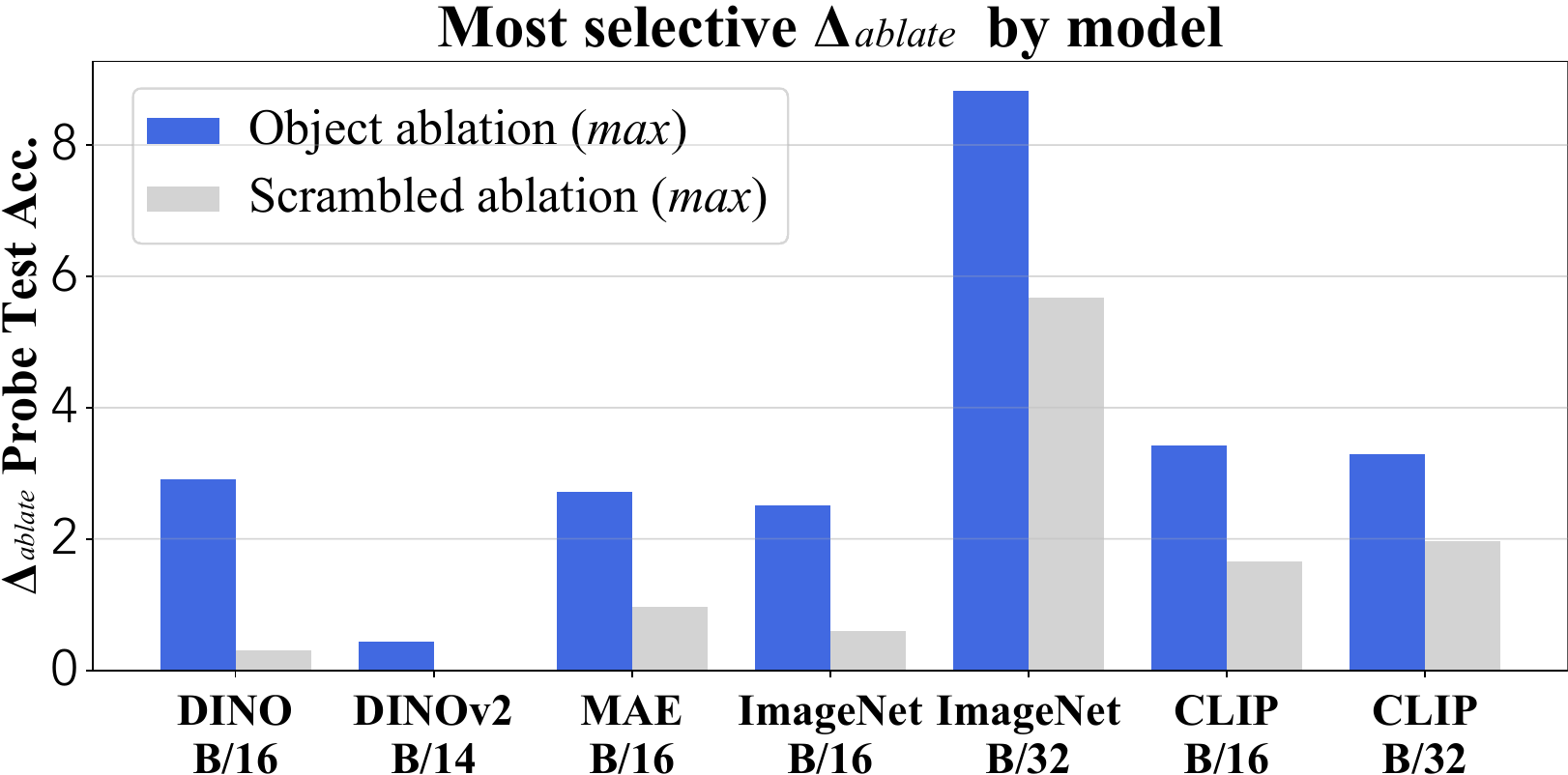}

   \caption{\textbf{Attention ablation results.} We plot the impact of ablating Gestalt continuity heads ($\Delta_{\text{\textit{ablate}}}$) on either the Object dataset or the Scrambled datasets. 
   We plot the most \textit{selective} ablations across layers for each model on the Blobs dataset, where selectivity is defined as  $\Delta_{\text{\textit{ablate}}}^\text{Object} - \Delta_{\text{\textit{ablate}}}^\text{Scrambled}$.
  We find that ablating Gestalt continuity heads impacts probe performance more on the Object dataset than the Scrambled datasets. See Appendix~\ref{appendix:ablations} for full results (including Curves results) and additional controls.}
   \label{fig:ablation}
\end{figure}

\vspace{-0.5em}
Having found that binding representations are sensitive to continuity in Section~\ref{sec:binding_probes}, and having identified a particular set of attention heads that are sensitive to continuity in Section~\ref{sec:gestalt_heads}, we now seek to understand whether Gestalt continuity heads play a role in the construction of 
binding representations.
To do so, we perform a mean ablation \citep{wanginterpretability} of the outputs of a set of Gestalt continuity heads and observe the impact on binding probe performance on the Object and Scrambled datasets. Specifically, for each probe, we mean-ablate the top 5 attention heads with the highest continuity sensitivity index scores (Section~\ref{sec:gestalt_heads}) across all layers preceding the probed representations.

In Figure~\ref{fig:ablation}, we see the results of this ablation analysis across models. We seek to demonstrate that these Gestalt continuity heads are \textit{selective}, or that ablating them impacts performance on the Object dataset more than the control datasets. Across models, we find that there exists a set of Gestalt continuity heads that are selectively implicated in producing binding vectors for the Object dataset. In Appendix~\ref{appendix:ablations} we compare the selectivity of the identified Gestalt continuity heads to other sets of randomly selected heads. We find that ablating the Gestalt continuity heads often impacts binding probe performance more selectively than  ablating random sets of heads.

Our earlier experiments explicitly demonstrated that binding representations are not uniquely the result of perceptual continuity cues, so we expect several heads beyond Gestalt continuity heads to be implicated in creating binding representations. However, our results suggest that Gestalt continuity heads play a role in the formation of object binding representations for many of the models.
\section{Discussion}
\label{sec:discussion}

\vspace{-0.5em}
\paragraph{Limitations and Future Work}
This work can be improved and extended in several ways. First, one can explore different ways to select Gestalt continuity heads for ablation. We rely on a simple heuristic for assessing selecting heads: taking the top-k heads according to continuity score, with $k=5$. More dynamic or layer-specific selection procedures might reveal even stronger contribution to binding.

Additionally, while our work reveals that continuity plays a role in producing binding representations, the degree to which it contributes when other Gestalt principles are available (as is the case with naturalistic stimuli) remains an open question.
Future work should try to discover attention heads that are sensitive to other Gestalt principles, such as closure or surface form similarity.
Finally, existing work has sought to understand the mechanisms behind ``detokenization'' in language models \citep{gurnee2023finding, feucht2024token}, or the process by which tokens are bound together to form meaningful units. Our work seeks to understand detokenization mechanisms in vision models by using principles from the vision sciences, but other work might apply techniques from language model mechanistic interpretability to this problem.

\vspace{-1em}
\paragraph{Conclusion}

This work employs a synthetic experimental paradigm and classic ideas from the vision sciences to investigate the mechanisms that give rise to object binding in vision transformers. We find that 1) continuity contributes to the formation of object binding representations (as discovered by binding probes), 2) specific heads in vision transformers track continuity across image patch boundaries, and 3) these heads oftentimes specifically contribute to the formation of the binding representation found by binding probes. In summary, this work advances our understanding of how contemporary vision models form object-level representations from low-level perceptual cues.

{
    \small
    \bibliographystyle{ieeenat_fullname}
    \bibliography{main}
}


\clearpage
\appendix
\begin{figure}[!ht]
  \centering
    \includegraphics[width=\linewidth]{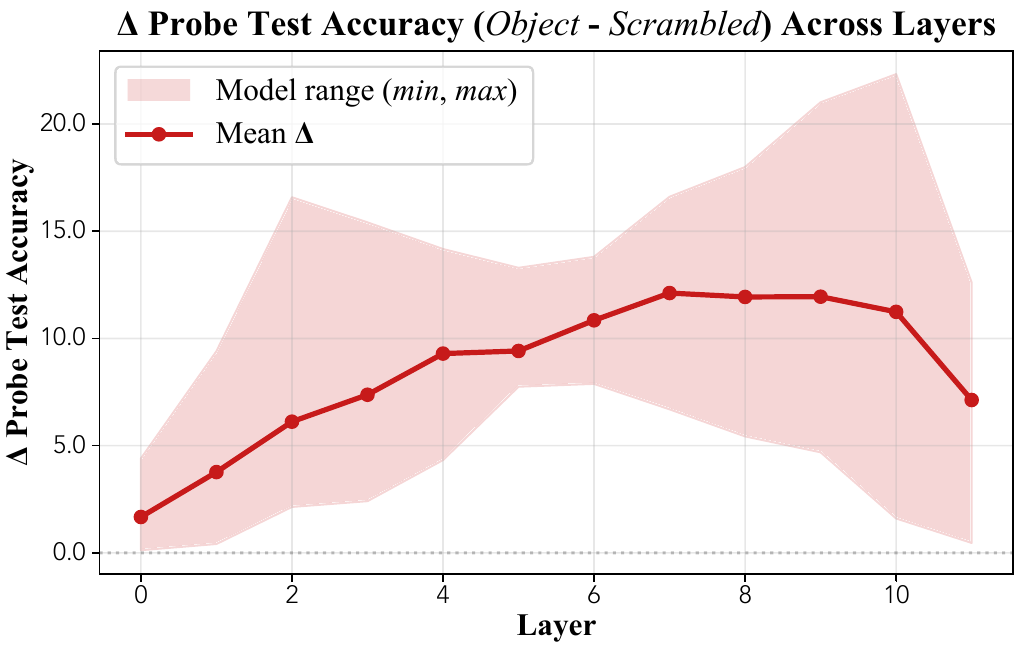}

   \caption{\textbf{Difference between object binding probe test accuracy on Curves Object stimuli vs. Scrambled stimuli.} These results follow the Blobs results shown in Figure~\ref{fig:probe}b.}
   \label{fig:probe_deltas_curves}
\end{figure}

\section{Dataset Generation \& Examples}
\label{appendix:datasets}

\begin{figure*}
  \centering
  \includegraphics[width=\linewidth]{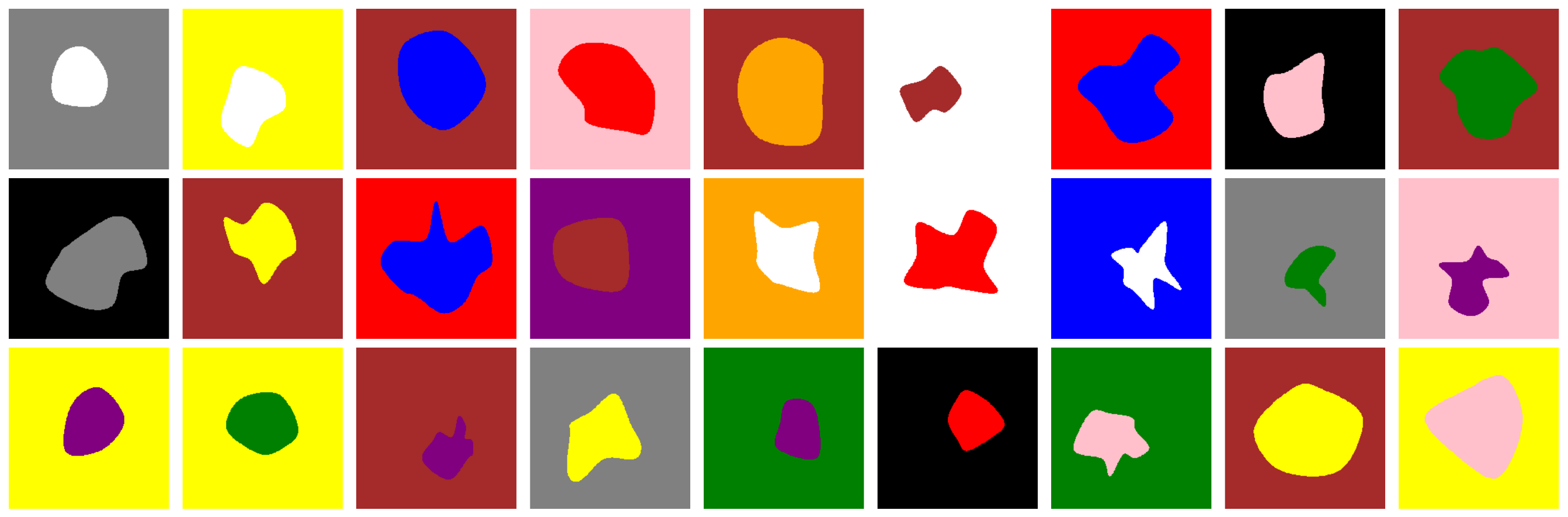}
  \caption{\textbf{Randomly selected stimuli from the Blobs dataset (``Object'' version).}}
  \label{fig:blob_gallery}
\end{figure*}
\begin{figure*}
  \centering
  \includegraphics[width=\linewidth]{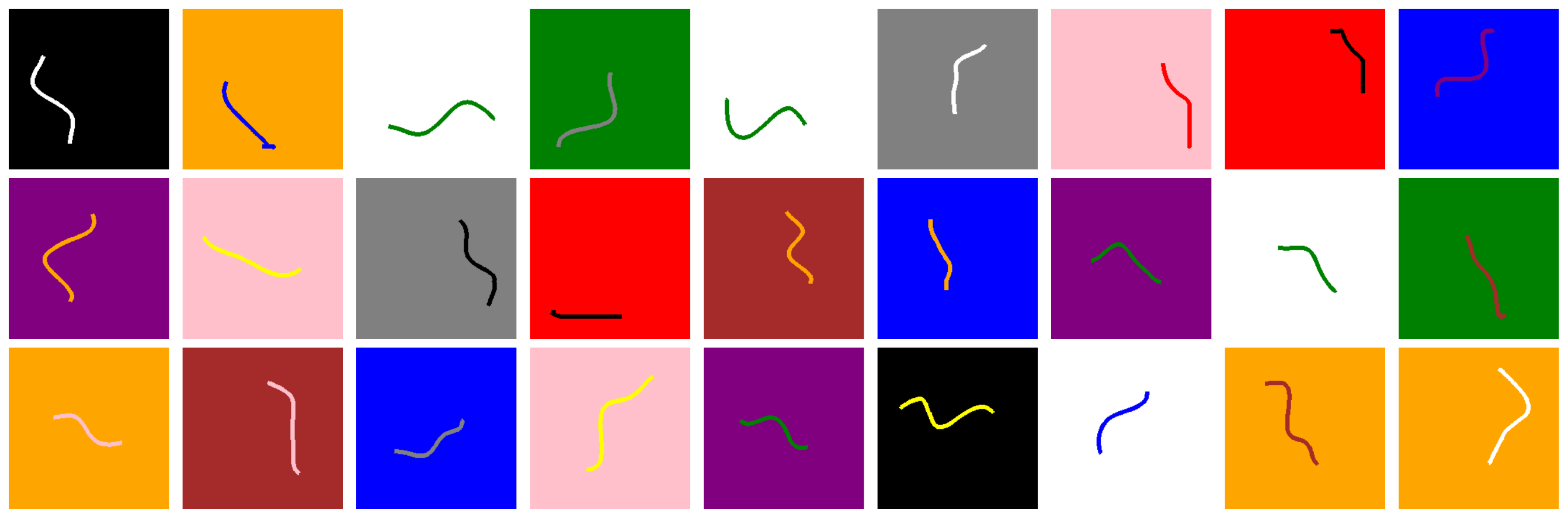}
  \caption{\textbf{Randomly selected stimuli from the Curves dataset (``Object'' version).}}
  \label{fig:curve_gallery}
\end{figure*}
\begin{figure*}
  \centering
  \includegraphics[width=\linewidth]{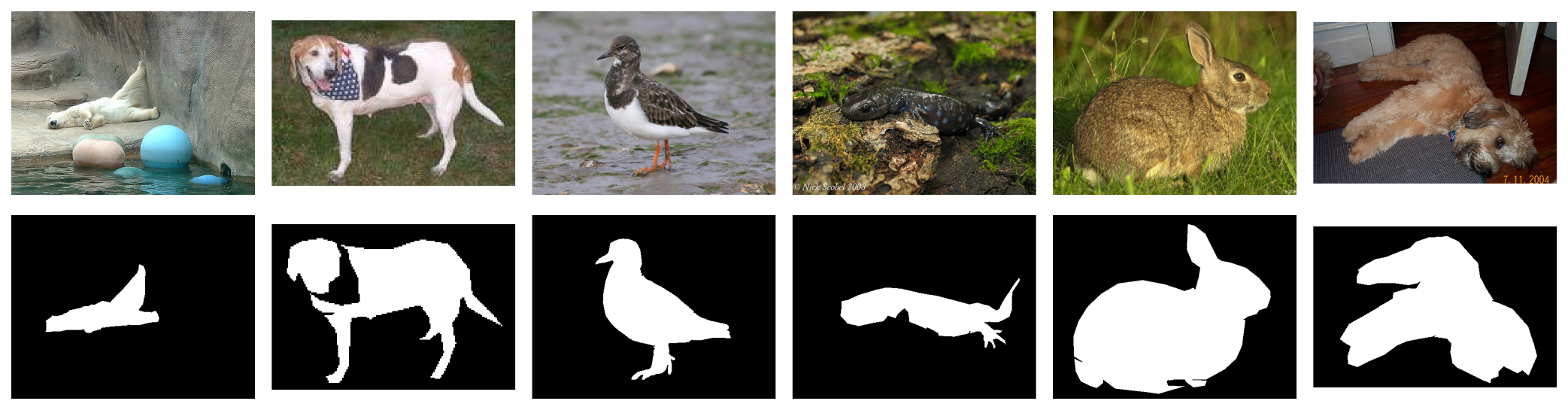}
  \caption{\textbf{Randomly selected stimuli from ImageNet (with corresponding object segmentations).}}
  \label{fig:imagenet_gallery}
\end{figure*}

Randomly selected images from each of the three datasets we use can be found in the following figures: Blobs (Figure~\ref{fig:blob_gallery}), Curves (Figure~\ref{fig:curve_gallery}), and ImageNet (Figure~\ref{fig:imagenet_gallery}).

Blobs (Figure~\ref{fig:blob_gallery}) are generated by first placing a shape center in a valid interior region of the image such that the blob radius (controlled by a size parameter) is fully within the image. Around this center, the contour is defined by a set of points distributed over radial angle with slight irregular spacing, each assigned a random distance from the center. The sampled points are interpolated into a smooth closed curve and filled, producing organic blob silhouettes with controllable complexity and substantial shape diversity. We vary blob sizes, complexity parameters, and colors from a predefined set of colors (red, blue, green, yellow, purple, orange, brown, pink, gray, black, white). The background color is selected from the same set minus the blob color.

Curves (Figure~\ref{fig:curve_gallery}) are generated the following procedure: a start point is placed inside a safe image window; subsequent candidate points are chosen with bounded step lengths and bounded turn angles (to encourage smooth continuation); and candidate points are iteratively regenerated if they leave the valid region, crowd existing points, or create self-intersections, yielding a non-looping open path with controlled complexity. The control points are  converted into a smooth visible stroke using spline interpolation, with a fallback straight segment if smoothing would collapse visually, producing continuous, organic open curves. Curves are 6 pixels thick, and their colors are sampled from the same colors as the Blobs dataset.

ImageNet (Figure~\ref{fig:imagenet_gallery}) stimuli and accompanying segmentations are taken from the test split of PartImageNet \cite{he2022partimagenet}. We use the segmentations to determine perimeter patches of objects. We only measure the continuity sensitivity index at $t=0$ for this dataset because translating or continuously rotating image patches creates unnatural discontinuities with the background. To create the control condition, we randomly rotate the target patch by $d\in\{90^\circ,270^\circ\}$ then select whichever rotation produces the maximal attention between neighboring patches.

\

\begin{figure*}
  \centering
  \includegraphics[width=\linewidth]{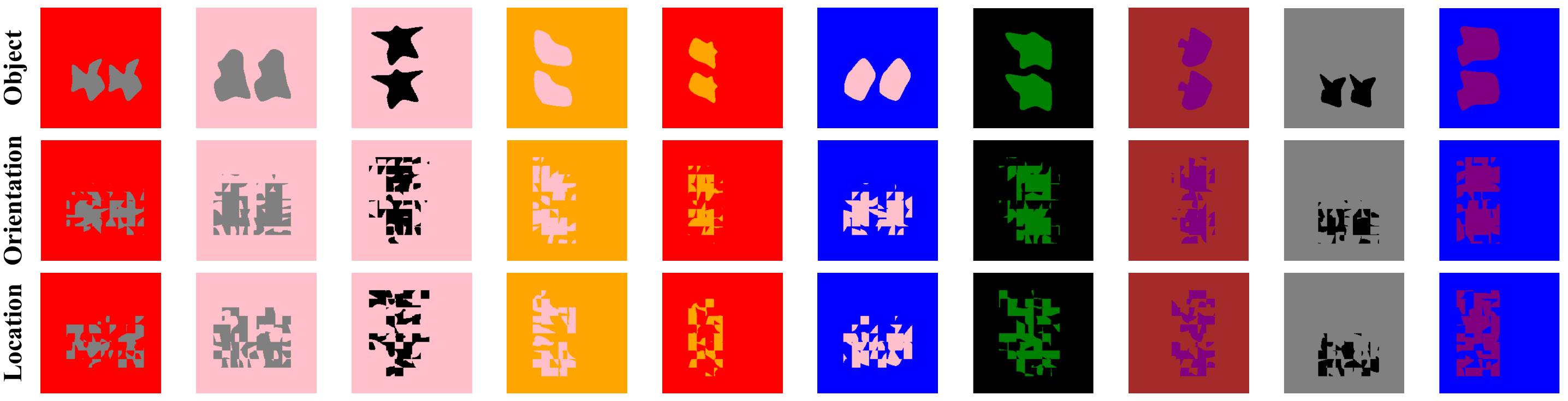}
  \caption{\textbf{Randomly selected triplets of stimuli showing the Object (top row), $\text{Scrambled}_{\textit{orientation}}$ (middle row), and $\text{Scrambled}_{\textit{location}}$ (bottom row) datasets for Blobs stimuli.} Specifically, these examples show Scrambled stimuli for the $16\times 16$ image patch size condition.}
  \label{fig:scrambled_gallery}
\end{figure*}

\noindent \textbf{Scrambled datasets}\qquad We generate $\text{Scrambled}_{\textit{orientation}}$ datasets for Blobs and Curves and for each model patch size (14, 16, 32) by randomly rotating each object-containing image patch by $d\in\{90^\circ,180^\circ,270^\circ\}$ degrees. We generate $\text{Scrambled}_{\textit{location}}$ datasets by first identifying the set of object-containing image patches, then randomly shuffling them within the set of positions they occupy. The patches are not rotated, but the shuffling disrupts the continuity of the shape boundary. See examples of both Scrambled datasets in Figure~\ref{fig:scrambled_gallery}. 

\

\begin{figure*}
  \centering
  \begin{subfigure}[t]{\linewidth}
    \includegraphics[width=\textwidth]{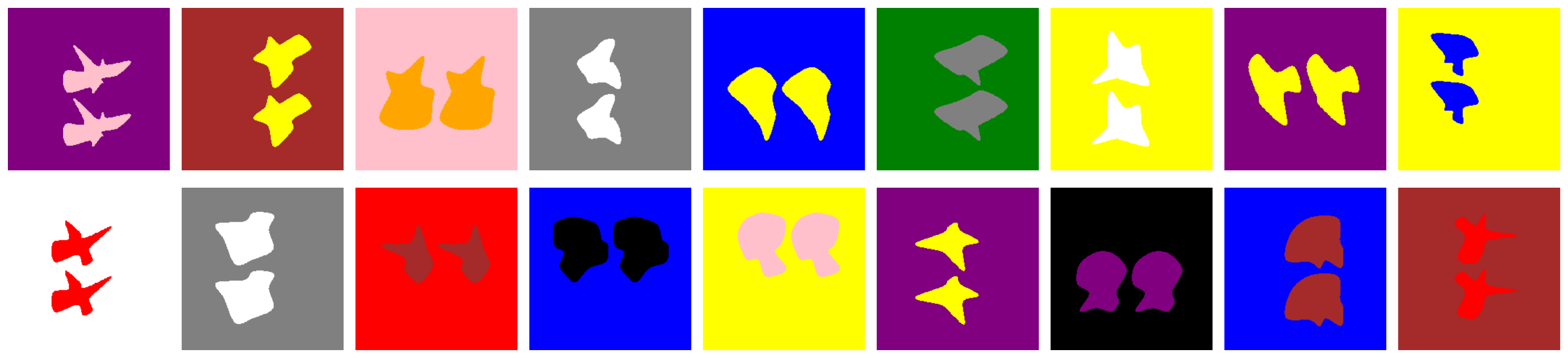}
    \caption{Examples from the Blobs binding dataset.}
  \end{subfigure}
  
  \begin{subfigure}[t]{\linewidth}
    \includegraphics[width=\textwidth]{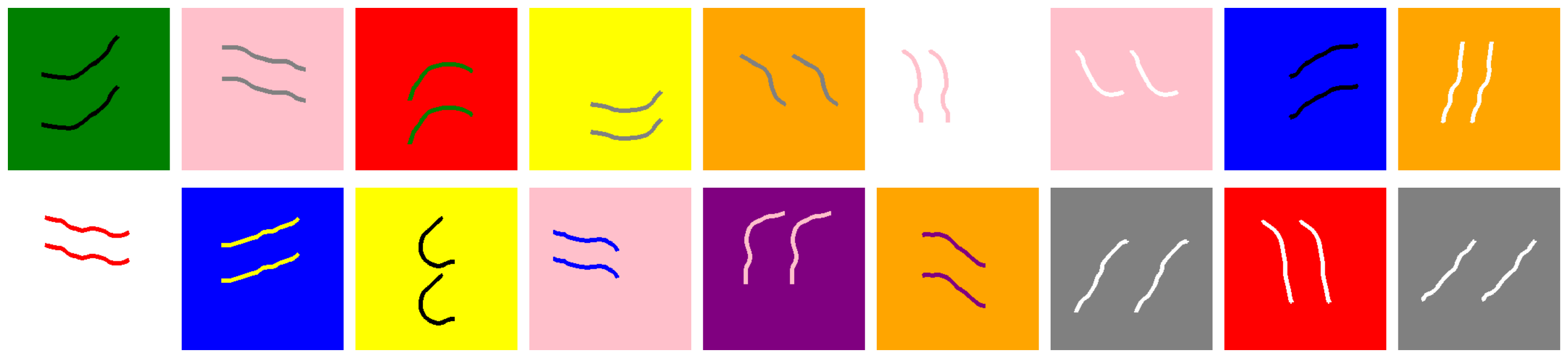}
    \caption{Examples from the Curves binding dataset.}
  \end{subfigure}
  \caption{\textbf{Randomly selected stimuli from the binding datasets used to train object binding probes in Section~\ref{sec:binding_probes}.} }
  \label{fig:binding_gallery}
\end{figure*}

\noindent \textbf{Binding datasets}\qquad Object binding datasets are generated exactly according to the Blobs and Curves dataset generation procedure described above. The sole difference is that instead of generating a single blob or curve, two copies of the identical blob or curve are placed in non-overlapping positions in the image. Blobs or curves are generated close together (maximum 12 pixels, minimum 5 pixels) so that the orientation and location scrambled datasets cannot be entirely solved by simple proximity heuristics (i.e. by detecting that patches nearby each other comprise the same object regardless of continuity cues). Examples of the Blobs and Curves binding datasets can be found in Figure~\ref{fig:binding_gallery}.

\section{Object Binding Probe Results}
\label{appendix:binding}

\begin{figure*}
  \centering
  \includegraphics[width=\linewidth]{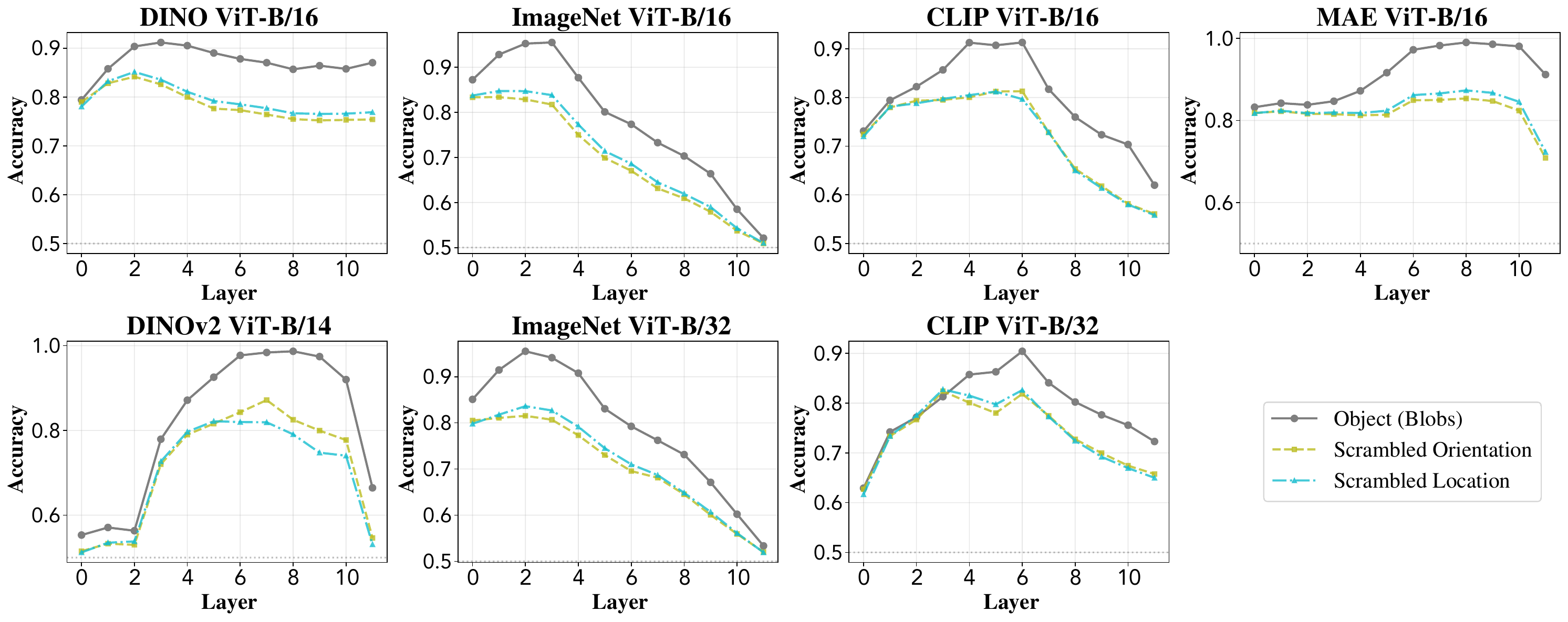}
  \caption{\textbf{Object binding probe test accuracy for Object, $\text{\textbf{Scrambled}}_{\text{\textit{orientation}}}$, and $\text{\textbf{Scrambled}}_{\text{\textit{location}}}$ datasets across layers for each model on the Blobs dataset.}}
  \label{fig:full_binding}
\end{figure*}

\begin{figure*}
  \centering
  \includegraphics[width=\linewidth]{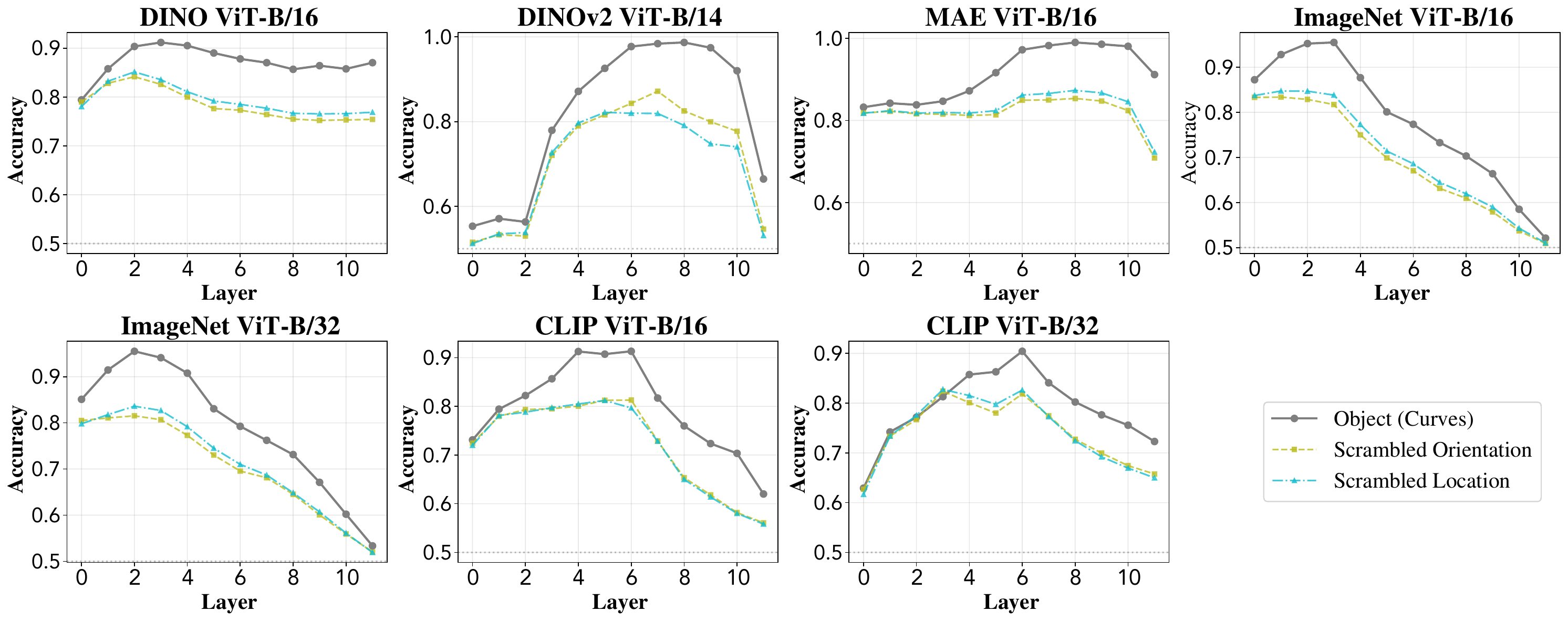}
  \caption{\textbf{Object binding probe test accuracy for Object, $\text{\textbf{Scrambled}}_{\text{\textit{orientation}}}$, and $\text{\textbf{Scrambled}}_{\text{\textit{location}}}$ datasets across layers for each model on the Curves dataset.}}
  \label{fig:full_binding_curves}
\end{figure*}

Object binding test probe accuracy for each of the three binding datasets across layers for each model can be found in Figure~\ref{fig:full_binding} (Blobs dataset) and Figure~\ref{fig:full_binding_curves} (Curves dataset). 
Figure~\ref{fig:probe_deltas_curves} shows the difference between object binding probe test accuracy on Curves Object stimuli vs.  Scrambled stimuli (taking the maximum probe test accuracy over $\{\text{Scrambled}_{\textit{orientation}}, \text{ Scrambled}_{\textit{location}}\}$), averaged across models (following Figure~\ref{fig:probe}b).
Maximum probe test accuracies for Object datasets vs. Scrambled datasets (following Figure~\ref{fig:probe}c) can be found in the tables in Figure~\ref{fig:object_vs_scrambled}. 

\begin{figure*}[t]
  \centering
  \begin{subfigure}[b]{0.45\linewidth}
    \centering
    \begin{tabular}{clc}
        & \multicolumn{2}{c}{$\max(\text{\textbf{Probe Test Acc.}})$} \\
        \toprule
        \textbf{Model} & \multicolumn{1}{c}{Object} & \multicolumn{1}{c}{Scrambled} \\
        \midrule
        DINO ViT-B/16 & 91.2\% \textcolor{royalblue}{(+6.1)} & 85.1\% \\
        DINOv2 ViT-B/14 & 98.7\% \textcolor{royalblue}{(+11.5)} & 87.2\% \\
        MAE ViT-B/16 & 99.0\% \textcolor{royalblue}{(+11.7)} & 87.3\% \\
        ImageNet ViT-B/16 & 95.5\% \textcolor{royalblue}{(+10.8)} & 84.7\% \\
        ImageNet ViT-B/32 & 95.5\% \textcolor{royalblue}{(+11.9)} & 83.6\% \\
        CLIP ViT-B/16 & 91.3\% \textcolor{royalblue}{(+10.0)} & 81.3\% \\
        CLIP ViT-B/32 & 90.4\% \textcolor{royalblue}{(+7.6)} & 82.8\% \\
        \bottomrule
    \end{tabular}
    \caption{Blobs dataset results.}
  \end{subfigure}
  \begin{subfigure}[b]{0.45\linewidth}
    \centering
    \begin{tabular}{clc}
        & \multicolumn{2}{c}{$\max(\text{\textbf{Probe Test Acc.}})$} \\
        \toprule
        \textbf{Model} & \multicolumn{1}{c}{Object} & \multicolumn{1}{c}{Scrambled} \\
        \midrule
        DINO ViT-B/16 & 86.9\% \textcolor{royalred}{(+10.6)} & 76.3\% \\
        DINOv2 ViT-B/14 & 97.3\% \textcolor{royalred}{(+16.6)} & 80.7\% \\
        MAE ViT-B/16 & 95.3\% \textcolor{royalred}{(+16.1)} & 79.2\% \\
        ImageNet ViT-B/16 & 89.4\% \textcolor{royalred}{(+11.7)} & 77.7\% \\
        ImageNet ViT-B/32 & 90.9\% \textcolor{royalred}{(+15.6)} & 75.3\% \\
        CLIP ViT-B/16 & 89.9\% \textcolor{royalred}{(+9.4)} & 80.5\% \\
        CLIP ViT-B/32 & 87.2\% \textcolor{royalred}{(+11.2)} & 76.0\% \\
        \bottomrule
    \end{tabular}
    \caption{Curves dataset results.}
  \end{subfigure}
  \caption{\textbf{Maximum object binding probe test accuracy on Object vs. Scrambled datasets.} These results are exactly the results shown in Figure~\ref{fig:probe}c, except the Scrambled results are shown explicitly. For the Scrambled results, we select the maximal probe test accuracy over $\{\text{Scrambled}_{\textit{orientation}}, \text{ Scrambled}_{\textit{location}}\}$.
  }
  \label{fig:object_vs_scrambled}
\end{figure*}
\section{Tuning Curve Examples}
\label{appendix:tuning_curves}

\begin{figure*}
  \centering
  \includegraphics[width=\linewidth]{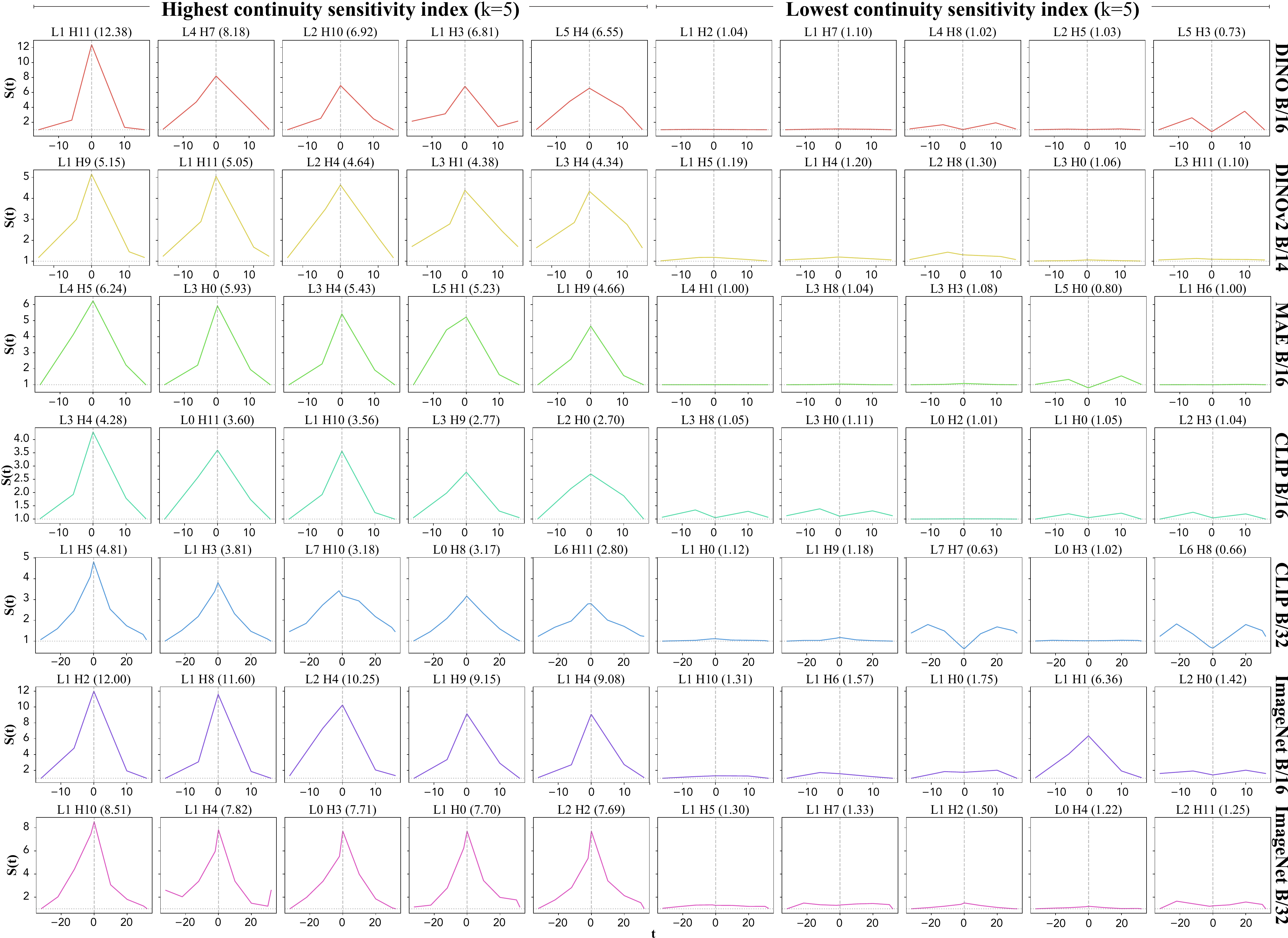}
  \caption{\textbf{Tuning curves for attention heads with the top-5 highest continuity sensitivity index scores (left panels) and the bottom-5 continuity sensitivity index scores (right panels) computed on the Blobs dataset for each model.} Tuning curves $S(t)$ are computed following the procedure described in Figure~\ref{fig:continuity_head}.; $S(t)$ essentially represents how many more times the attention head is sensitive to the curve-completing condition vs. the non curve-completing control, thereby giving a measure of sensitivity to continuity across patches (as opposed to mere pixel-level similarity or proximity). The $x$-axis shows $t$, i.e. the amount of displacement of the moving image patch (in pixels) from its original, maximally-aligned position. The $y$-axis shows the continuity sensitivity index, $S(t)$. The timestep $t=0$ gives the scores shown in the heatmaps in Figure~\ref{fig:heatmaps_blobs}. Above each tuning curve is the layer (L) and head (H) index, followed by $S(0)$ in parentheses.}
  \label{fig:tuning_curves}
\end{figure*}

Additional tuning curves computed on the Blobs dataset (such as the curve $S(t)$ plotted in Figure~\ref{fig:continuity_head}c) are shown in Figure~\ref{fig:tuning_curves} for each model. The 5 leftmost panels show heads with the highest continuity sensitivity index, while the rightmost panels show heads with the lowest continuity sensitivity index. The heads with the highest scores show the characteristic shape expected of attention heads that are sensitive to Gestalt continuity; on the other hand, the heads with the lowest scores do not show this pattern. 
 
\section{Continuity Sensitivity Index at $t=0$ for  All Heads, Layers, and Models}
\label{appendix:continuity}

\begin{figure*}
  \centering
  \includegraphics[width=\linewidth]{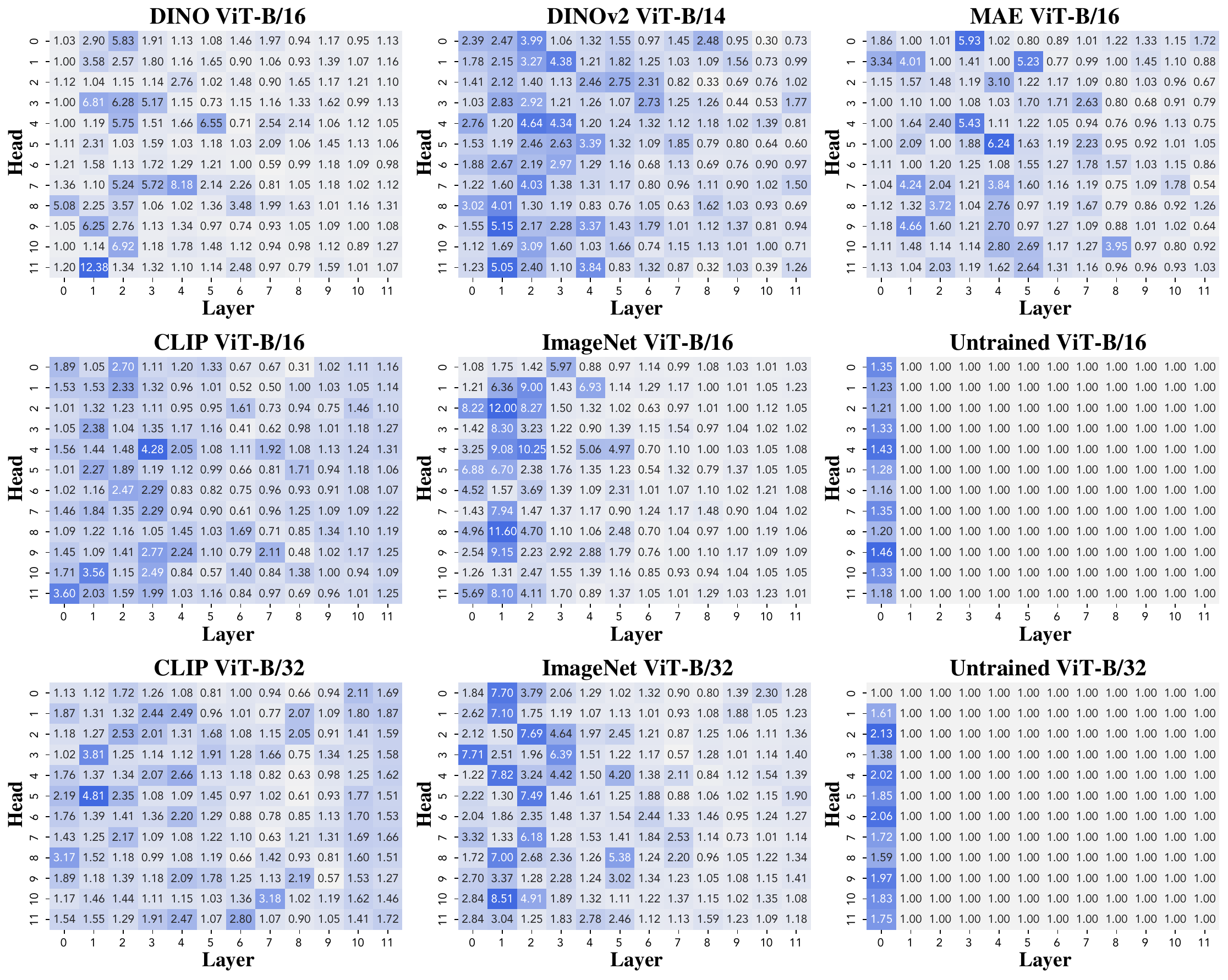}
  \caption{\textbf{Continuity sensitivity index scores computed on the Blobs dataset at $t=0$ for all heads, layers, and models.} Examples of stimuli from the Blobs dataset are shown in Figure~\ref{fig:blob_gallery}. Heads with the highest continuity scores tend to be clustered within the first four model layers.}
  \label{fig:heatmaps_blobs}
\end{figure*}

\begin{figure*}
  \centering
  \includegraphics[width=\linewidth]{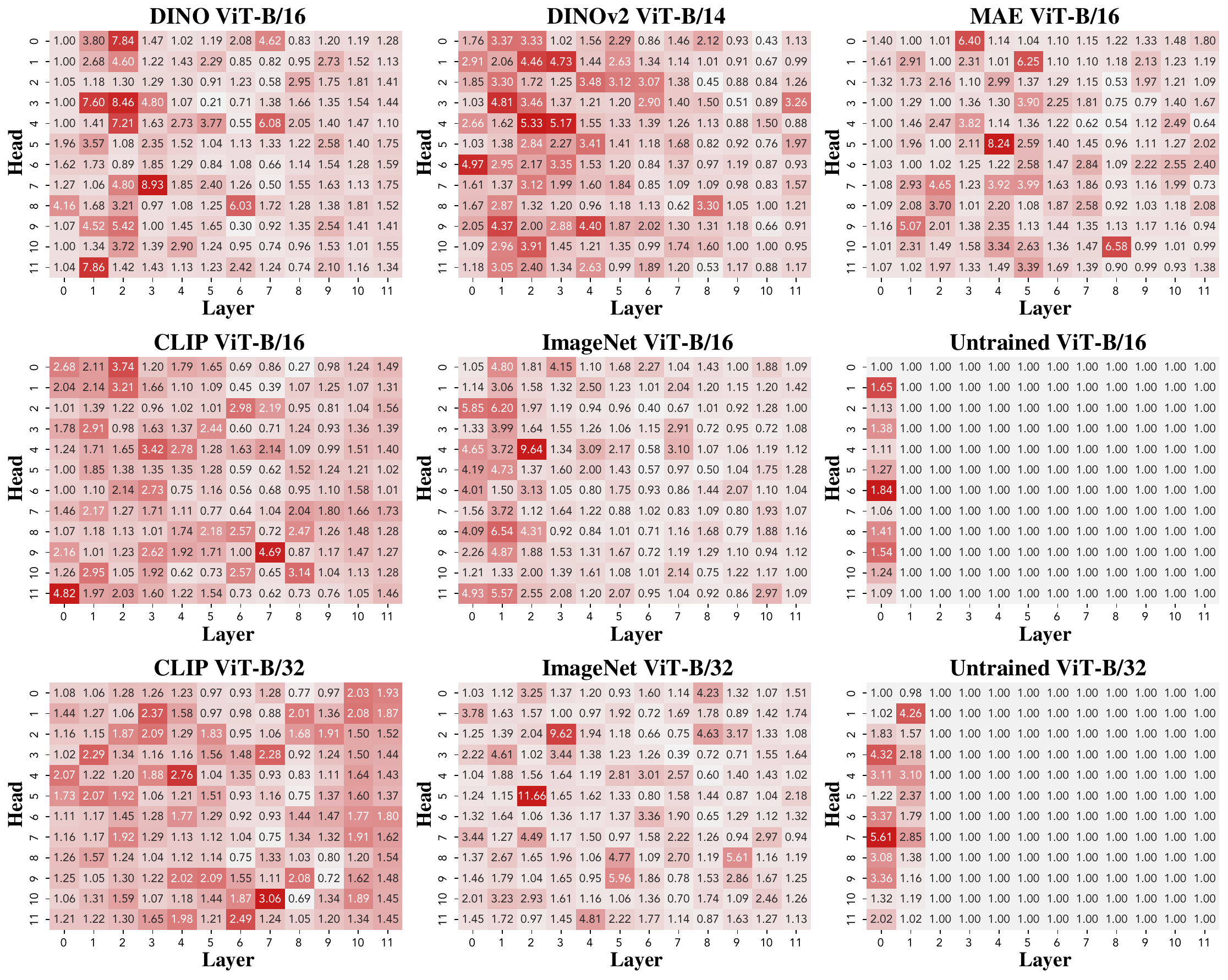}
  \caption{\textbf{Continuity sensitivity index scores computed on the Curves dataset at $t=0$ for all heads, layers, and models.} Examples of stimuli from the Curves dataset are shown in Figure~\ref{fig:curve_gallery}.}
  \label{fig:heatmaps_curves}
\end{figure*}

\begin{figure*}
  \centering
  \includegraphics[width=\linewidth]{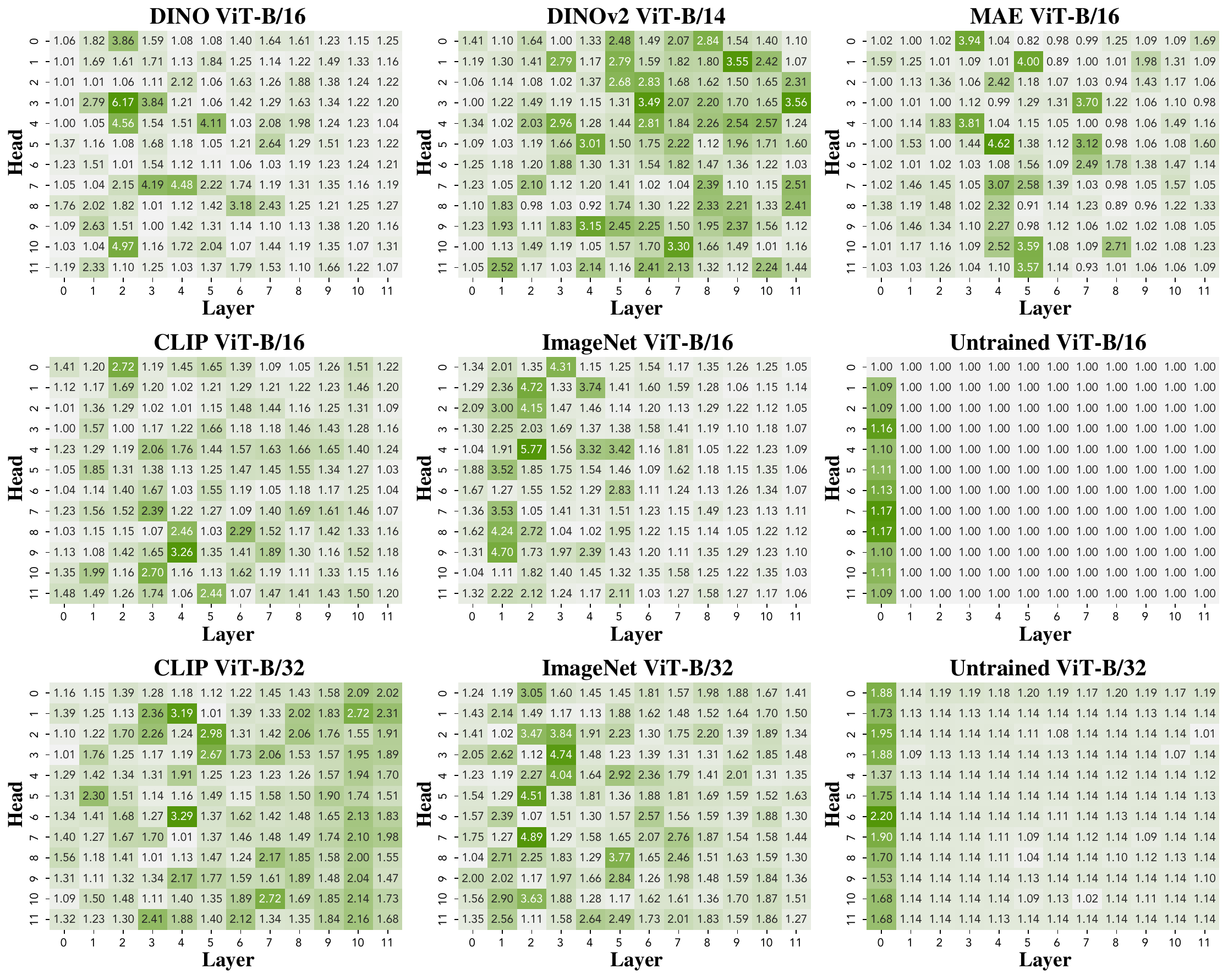}
  \caption{\textbf{Continuity sensitivity index scores computed on the ImageNet dataset at $t=0$ for all heads, layers, and models.} Examples of stimuli from the ImageNet dataset are shown in Figure~\ref{fig:imagenet_gallery}.}
  \label{fig:heatmaps_imagenet}
\end{figure*}

We compute continuity scores (i.e. continuity sensitivity index at $t=0$) for each attention head in each layer for each model. These scores are displayed in the heatmaps in Figure~\ref{fig:heatmaps_blobs} (Blobs dataset), Figure~\ref{fig:heatmaps_curves} (Curves), and Figure~\ref{fig:heatmaps_imagenet} (ImageNet). We also include results for two untrained models which have an identical architecture to the ImageNet ViTs (all models have highly similar architectures). Untrained models have scores of $1$ everywhere (meaning no difference between continuous patches and non-continuous patches) except the $0$th layer; this is likely because the $0$th layer represents pixel-level information from the image and is likely not attributable to any representational competence of the untrained models.

We note that the concentration of continuity heads across layers roughly tracks probe performance across layers (see Appendix~\ref{appendix:binding}). In particular, DINOv2 ViT-B/14 and MAE ViT-B/16 have continuity heads through layers 4 and 5, and probe performance rises at these layers. On the other hand, ImageNet models tend to have continuity heads concentrated in layers 1 and 2, and probe performance also peaks around those layers. DINO ViT-B/16 has its greatest number of continuity heads in layer 2, and probe performance rises by layer 2. CLIP ViT-B/16 has its continuity heads concentrated before layer 4, which coincides with its peak in probe performance. CLIP ViT-B/32 has some later continuity heads, and correspondingly peaks in probe accuracy later.

\section{Correlation of Continuity Heads Across Datasets}
\label{appendix:correlations}
\begin{figure*}
  \centering
  \includegraphics[width=\linewidth]{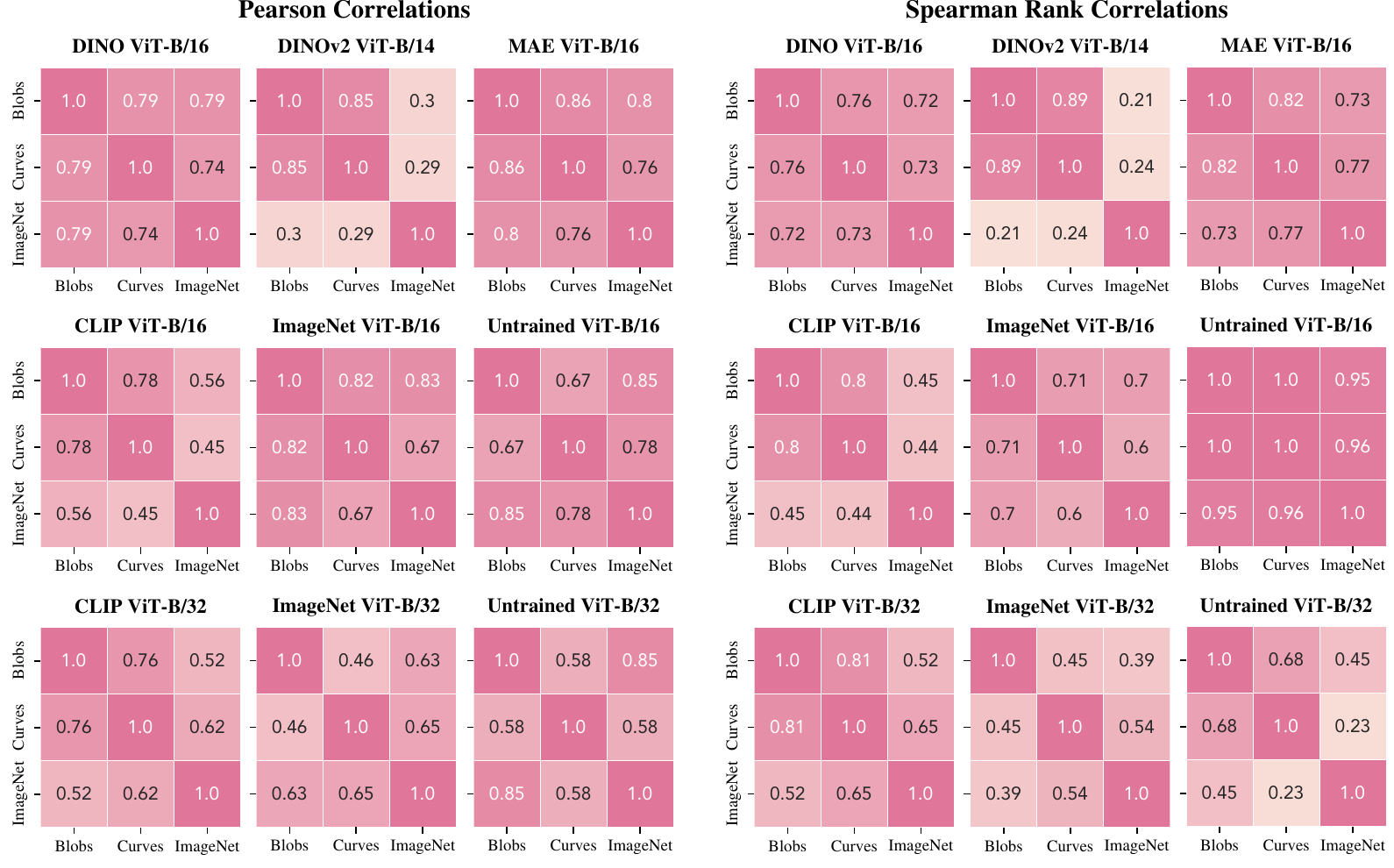}
  \caption{\textbf{Pearson's $r$ (left) and Spearman's $\rho$ for continuity sensitivity index scores at $t=0$ between Blobs, Curves, and ImageNet datasets.} See Appendix~\ref{appendix:datasets} for details of how the $t=0$ stimuli are generated for each dataset.}
  \label{fig:correlations}
\end{figure*}

To test whether continuity heads are robust across datasets, we compute Pearson's $r$ and Spearman's $\rho$ of the continuity sensitivity index scores at $t=0$ across three datasets: our main Blobs dataset, our Curves dataset, and ImageNet objects. Specifically, we compute the pairwise correlations between the length $144$ vector of continuity scores for each model ($12$ layers $\times$ $12$ attention heads per layer). Figure~\ref{fig:correlations} contains the $r$ and $\rho$ for each model and dataset pair. 

For Pearson's $r$, $p$-values range from $1.43e-42$ to $5.05e-04$. For Spearman's $\rho$, $p$-values range from $3.20e-156$ to $1.06e-02$.

\section{Causal Ablation Results}
\label{appendix:ablations}

\begin{figure*}
  \centering
  \includegraphics[width=\linewidth]{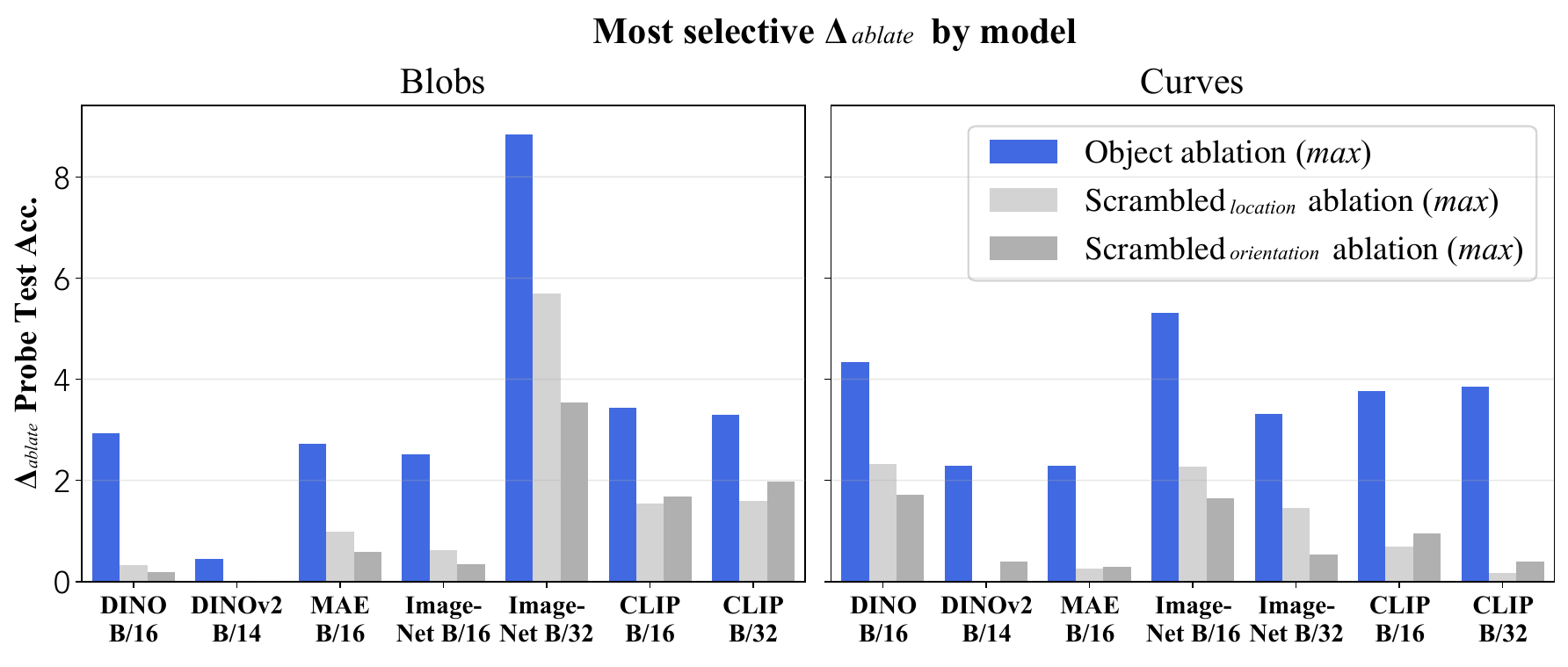}
  \caption{\textbf{Attention ablation results for Blobs (left) and Curves (right) datasets.} As in Figure~\ref{fig:ablation}, each bar shows the $\Delta_{ablate}$ (i.e. decrement in object binding probe test accuracy as a result of the ablation). We visualize the most selective ablation per model (i.e., the layer at which ablation impacts the Object dataset more than the Scrambled datasets). While Figure~\ref{fig:ablation} only shows the maximum control ablation (taken across both Scrambled datasets), here we show results for both Scrambled datasets separately.}
  \label{fig:ablation_all}
\end{figure*}

Results following Figure~\ref{fig:ablation} for both Blobs and Curves datasets, broken out by each Scrambled dataset, can be found in Figure~\ref{fig:ablation_all}. 

For each model and at each model layer, we compute the impact of attention ablations on probe test accuracy for continuity ablations. As described in Section~\ref{sec:ablation}, continuity ablations are performed as follows: for a given model layer, mean-ablate the top five attention heads (with the highest continuity scores from Section~\ref{sec:gestalt_heads}) across all previous layers; then, measure the impact on the given layer's object probe test accuracy (using images from the Object versions of each dataset --- that is, on intact blobs and curves). 

\begin{figure*}
  \centering
  \includegraphics[width=\linewidth]{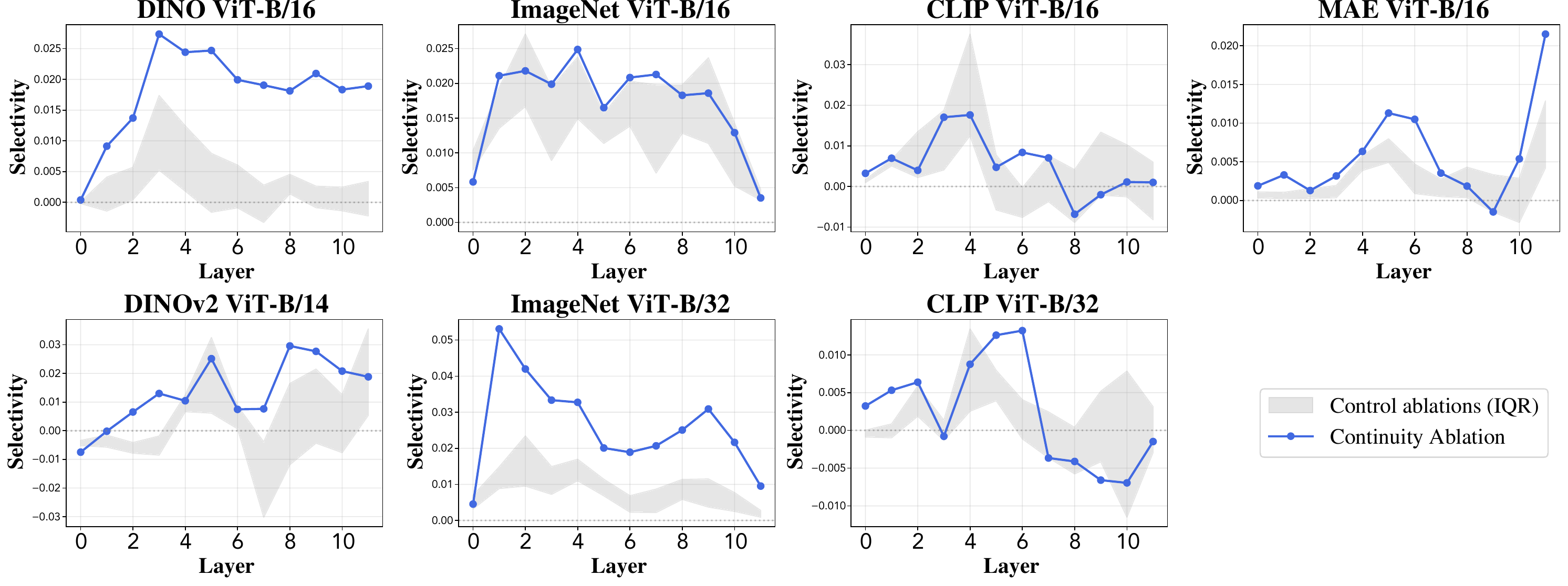}
  \caption{\textbf{Selectivity (i.e. $\Delta_{ablate} \text{Object}-\Delta_{ablate} \text{Scrambled}$) for the $\text{Scrambled}_{\textit{orientation}}$ dataset (Blobs).} Here, the continuity heads are ablated on the Object version of the dataset to compute $\Delta_{ablate} \text{Object}$; the same set of heads are ablated on the $\text{Scrambled}_{\textit{orientation}}$ dataset to compute $\Delta_{ablate} \text{Scrambled}$.}
  \label{fig:causal_blobs_orientation}
\end{figure*}

\begin{figure*}
  \centering
  \includegraphics[width=\linewidth]{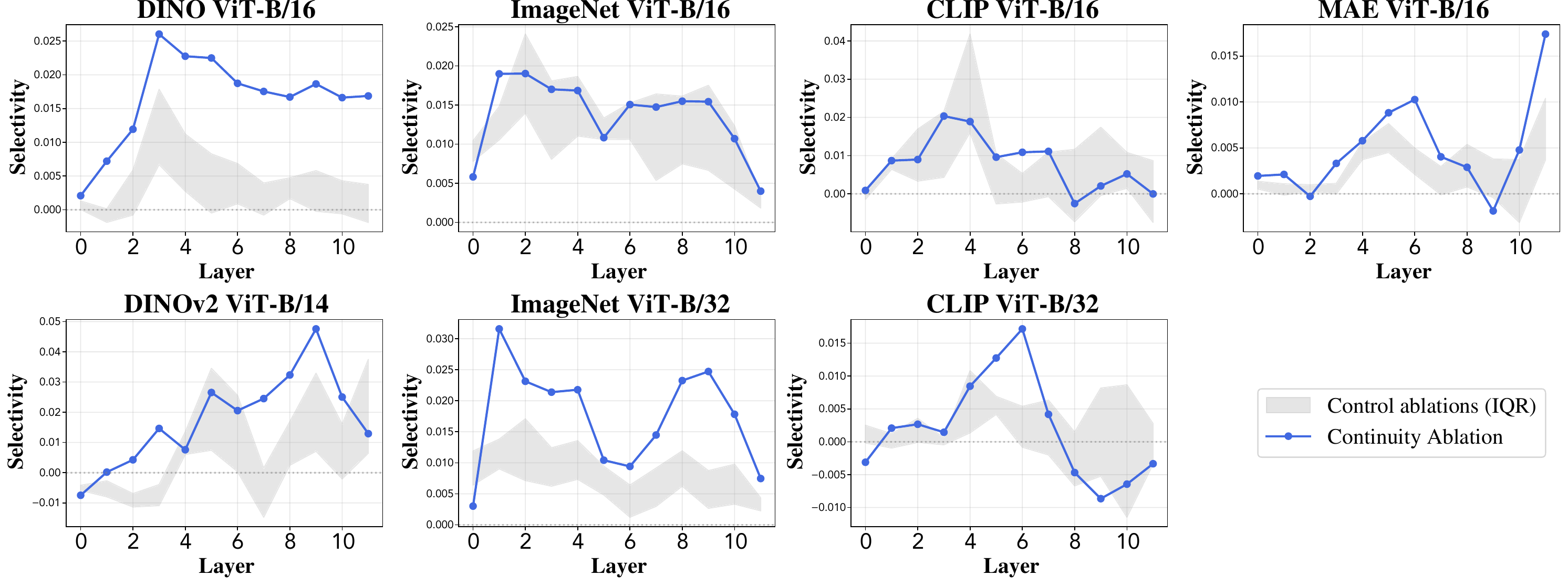}
  \caption{\textbf{Selectivity (i.e. $\Delta_{ablate} \text{Object}-\Delta_{ablate} \text{Scrambled}$) for the $\text{Scrambled}_{\textit{location}}$ dataset (Blobs).} Here, the continuity heads are ablated on the Object version of the dataset to compute $\Delta_{ablate} \text{Object}$; the same set of heads are ablated on the $\text{Scrambled}_{\textit{location}}$ dataset to compute $\Delta_{ablate} \text{Scrambled}$.}
  \label{fig:causal_blobs_location}
\end{figure*}

\begin{figure*}
  \centering
  \includegraphics[width=\linewidth]{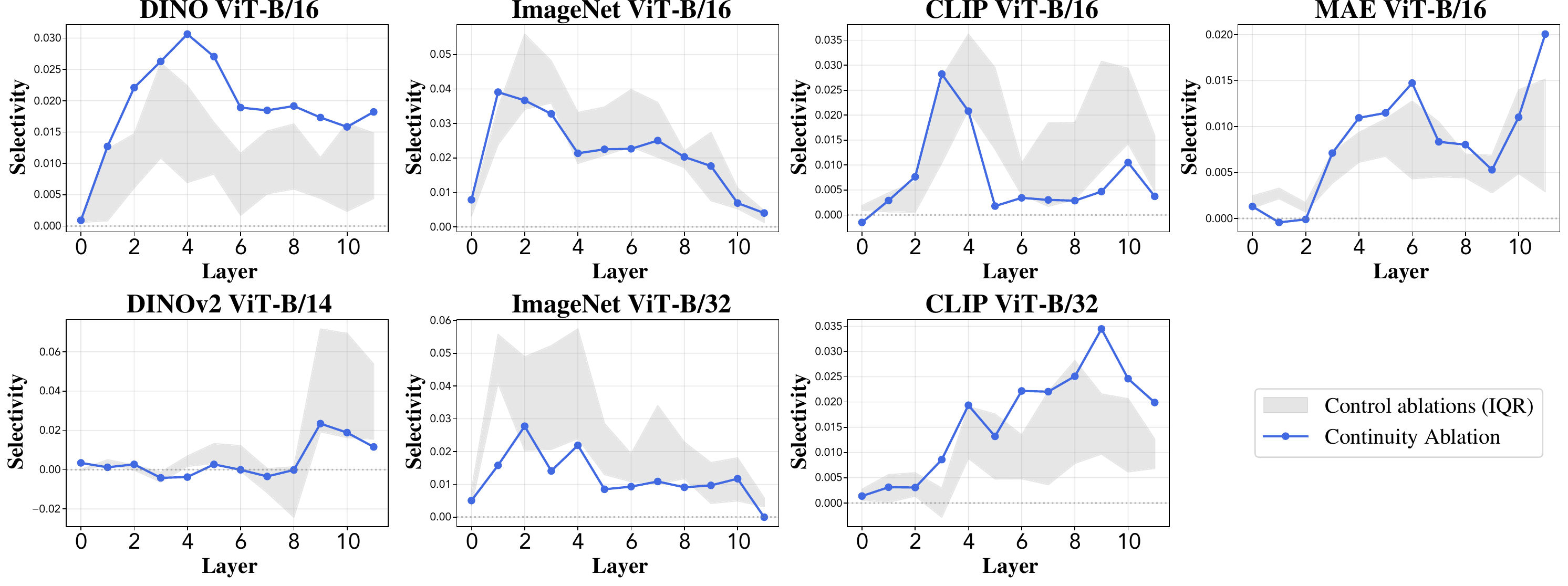}
  \caption{\textbf{Selectivity (i.e. $\Delta_{ablate} \text{Object}-\Delta_{ablate} \text{Scrambled}$) for the $\text{Scrambled}_{\textit{orientation}}$ dataset (Curves).} Here, the continuity heads are ablated on the Object version of the dataset to compute $\Delta_{ablate} \text{Object}$; the same set of heads are ablated on the $\text{Scrambled}_{\textit{orientation}}$ dataset to compute $\Delta_{ablate} \text{Scrambled}$.}
  \label{fig:causal_curves_orientation}
\end{figure*}

\begin{figure*}
  \centering
  \includegraphics[width=\linewidth]{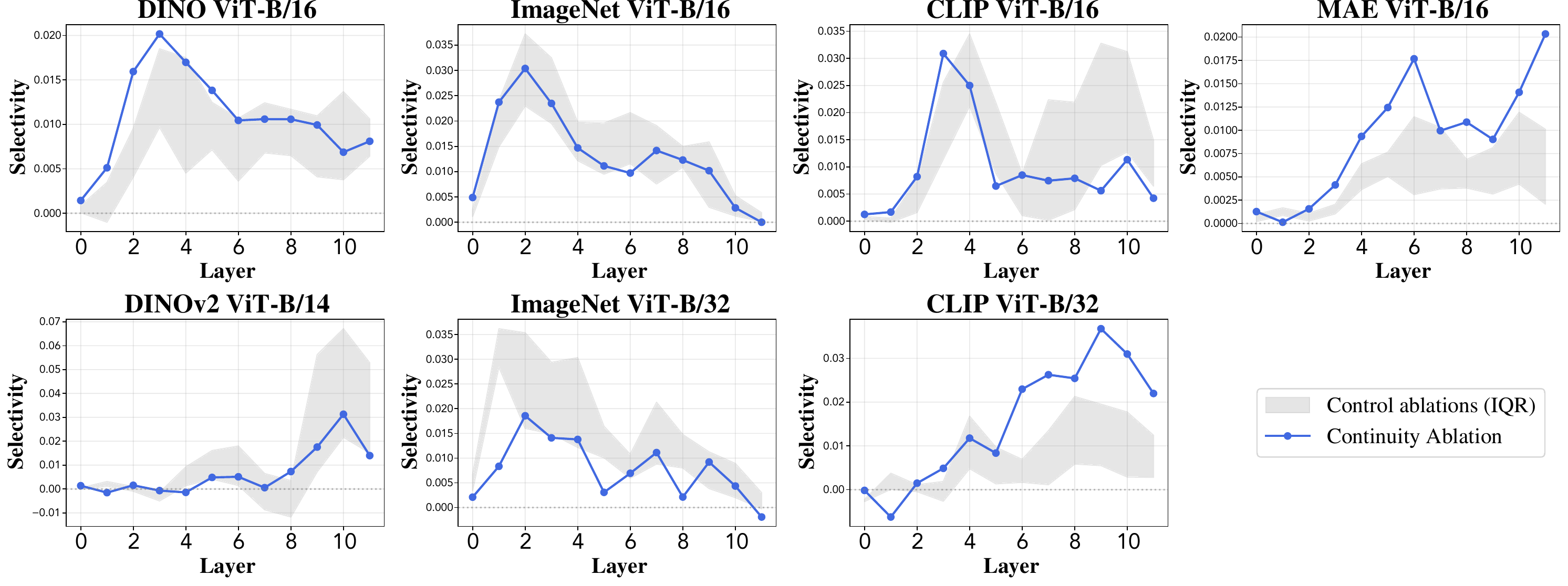}
  \caption{\textbf{Selectivity (i.e. $\Delta_{ablate} \text{Object}-\Delta_{ablate} \text{Scrambled}$) for the $\text{Scrambled}_{\textit{location}}$ dataset (Curves).} Here, the continuity heads are ablated on the Object version of the dataset to compute $\Delta_{ablate} \text{Object}$; the same set of heads are ablated on the $\text{Scrambled}_{\textit{location}}$ dataset to compute $\Delta_{ablate} \text{Scrambled}$.}
  \label{fig:causal_curves_location}
\end{figure*}

We first compare the effect of continuity ablations to the effect ablating the same heads on our control datasets: $\text{Scrambled}_{\textit{orientation}}$ and $\text{ Scrambled}_{\textit{location}}$. Specifically, we compute a quantity called \emph{Selectivity}, which is defined as follows. Let $\Delta_{ablate}^\text{Object}$ represent the decrement in probe test accuracy on the Object dataset as a result of mean-ablating the continuity heads, and $\Delta_{ablate}^\text{Scrambled}$ represent the decrement in probe test accuracy on either of the Scrambled datasets  as a result of ablating the same set of heads. Then $\text{Selectivity}=\Delta_{ablate} \text{Object}-\Delta_{ablate} \text{Scrambled}$, i.e. how much more the Object binding probe is impacted by ablating continuity heads vs. the Scrambled binding probes. Selectivity for the Blobs dataset is shown in Figure~\ref{fig:causal_blobs_orientation} ($\text{Scrambled}_{\textit{orientation}}$) and Figure~\ref{fig:causal_blobs_location} ($\text{ Scrambled}_{\textit{location}}$). Selectivity for the Curves dataset is shown in Figure~\ref{fig:causal_curves_orientation} ($\text{Scrambled}_{\textit{orientation}}$) and Figure~\ref{fig:causal_curves_location} ($\text{ Scrambled}_{\textit{location}}$). 

We compare the selectivity of our set of Gestalt continuity heads to sets of randomly selected heads. 20 control sets (i.e. random head selections) are performed for each model layer. For each of these random sets, we ensure that the same number of heads are ablated per layer as in the set of Gestalt continuity heads. For example, if the Gestalt continuity heads ablated when assessing probe performance at layer 4 include 2 heads at layer 2, 1 at layer 3 and 2 at layer 4, then every one of the control sets will have 2 heads from layer 2, 1 at layer 3, and 2 at layer 4. We visualize the interquartile range of control set selectivities in Figures~\ref{fig:causal_blobs_orientation}-~\ref{fig:causal_curves_location}. We find that the Gestalt continuity head ablations are typically more selective than the control ablations, at least for 1 layer per model. 

Note that our method for selecting Gestalt continuity heads to ablate is quite crude. Many models have far more than 5 heads that could plausibly be called Gestalt continuity heads, and so we expect that some control ablations include these. Given this caveat, we view these ablation results as positive, but somewhat preliminary.


\end{document}